\documentclass{article}

\usepackage[numbers]{natbib}

\usepackage[final]{neurips_2020_ml4ad}

\usepackage[utf8]{inputenc} 
\usepackage[T1]{fontenc}    
\usepackage{hyperref}       
\usepackage{url}            
\usepackage{booktabs}       
\usepackage{amsfonts}       
\usepackage{nicefrac}       
\usepackage{microtype}      
\usepackage{graphicx}
\usepackage{multirow}
\usepackage{caption}
\usepackage{subcaption}
\title{Single Shot Multitask Pedestrian Detection and Behavior Prediction}

\author{%
  Prateek Agrawal \\
  Volkswagen Group\\
  Innovation Center California\\
  Belmont, California 94002\\
  \texttt{prateek.agrawal@vw.com} \\
  \And
  Pratik Prabhanjan Brahma \\
  Volkswagen Group\\
  Innovation Center California \\
  Belmont, California \\
  \texttt{pratik.brahma@vw.com} \\
}

\begin{document}

\graphicspath{ {./images/} }

\maketitle

\begin{abstract}
Detecting and predicting the behavior of pedestrians is extremely crucial for self-driving vehicles to plan and interact with them safely. Although there have been several research works in this area, it is important to have fast and memory efficient models such that it can operate in embedded hardware in these autonomous machines. In this work, we propose a novel architecture using spatial-temporal multi-tasking to do camera based pedestrian detection and intention prediction. Our approach significantly reduces the latency by being able to detect and predict all pedestrians' intention in a single shot manner while also being able to attain better accuracy by sharing features with relevant object level information and interactions. 

\end{abstract}

\section{Introduction}

A typical autonomous driving modular stack consists of several modules like sensors, perception, localization, prediction, planning, and control. The perception module consists of tasks, like object detection or free space detection, that perceive the environment around the ego vehicle. The prediction module can anticipate the intended behavior and future trajectories of traffic agents. In order to deal safely with Vulnerable Road Users (VRUs) like pedestrians, both of these modules are extremely crucial. 



The standard pipeline typically perceives the pedestrians first. This is followed by predicting their behavior or future trajectory by using information like their past motion from corresponding sensor data. We call this as the sequential approach. Pedestrians typically do not follow kinematic models like vehicles and thus it is challenging to have fast and highly accurate predictions in a computationally efficient manner. As an overall system, the sequential approach demands high memory usage and high inference time which may be problematic given the limited computing capability.

In this work, we present a novel approach to solve the problem of pedestrian intention prediction, as defined in \cite{Rasouli2019PIE}, by taking a parallel multi-tasking approach between detection and prediction. As shown in Figure 1, we hypothesize that our approach can have manifold advantages: (a) Low latency by executing parts of prediction and perception in parallel, (b) Practical and fixed inference time by detecting and predicting for all pedestrians in a single shot manner, (c) Low parameter and feature memory usage, (d) Multi-tasking leverages correlations to help improving accuracy.
In the following sections, we explain the design and compare
its performance with a baseline sequential approach on the Pedestrian Intention Estimation (PIE) data set \cite{Rasouli2019PIE}.

\begin{figure}[h]
  \centering
  \includegraphics[scale=0.6]{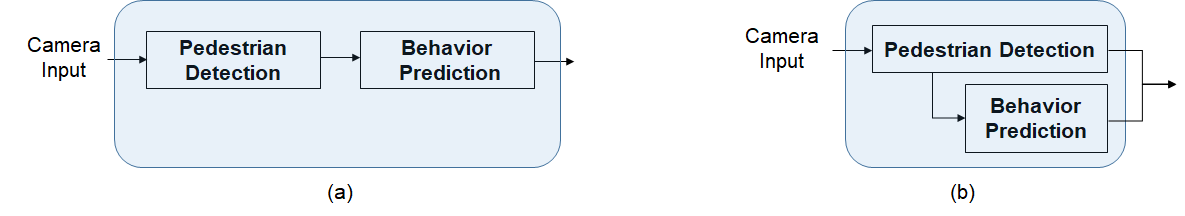}
  \caption{(a) Sequential approach with perception followed by prediction, and (b) proposed auxiliary approach that multi-tasks both followed by a spatial association of pedestrian  behavior}
\end{figure}

\section{Related Work}
\label{headings}

\paragraph{Pedestrian understanding} Several recent works \cite{liu2019center, liu2019high, hasan2020pedestrian} have pushed the boundaries in the field of pedestrian detection. Also for most multi-object detection and pixel wise semantic segmentation driving data sets \cite{cordts2016cityscapes, yu2020bdd100k}, pedestrian is usually a distinct category. Pedestrian understanding however goes beyond that by attempting to detect multiple aspects \cite{camara2020pedestrian} like pose \cite{fang2019intention}, gesture \cite{rautaray2015vision} and actions \cite{chen2020survey} of human beings and being able to predict the intended behavior and eventually the actual trajectory that the pedestrian is expected to execute in future. Most of the approaches \cite{Rasouli2019PIE, lorenzo2020rnn} for intention prediction however fall under the sequential approach mentioned in Figure 1(a). Predicting pedestrian behavior is also an interactive problem since each person's future depends on the state of other VRUs, vehicles, traffic lights and other proximal map elements. Although there have been several works \cite{gupta2018social, thiede2019analyzing, alahi2016social} towards conducting a social prediction for all interacting agents, most of these again assume that the pedestrians have already been detected and are being tracked.



\paragraph{Multi-task learning}Many works \cite{ruder2017overview, misra2016cross, kendall2018multi} have shown the advantage of using multi-task learning for different applications, like Mask R-CNN \cite{he2017mask} for instance level segmentation. There have been efforts like \cite{pop2019multi} to do multi-task pedestrian detection and action recognition but the actual time-to-cross prediction is only done after the bounding boxes are obtained through detection. It is also similar in \cite{ranga2020vrunet} where multiple prediction sub-modules are multi-tasked but only run after obtaining semantic segmentation output. PnPNet \cite{liang2020pnpnet} proposes end-to-end learning of perception and prediction but the architecture still executes the trajectory prediction LSTM only after the actual detection network. FaF \cite{luo2018fast} is another end-to-end solution for detection, tracking and motion forecasting but also places the forecasting prediction model after detection. Thus, the inference time for these methods is typically large. Contrary to these approaches, our proposed methodology leverages low level feature sharing to capture better interactive information and parallel execution of task specific layers for doing detection and behavior prediction of all pedestrians simultaneously.    


\section{Methodology}
\label{others}

The baseline model that we compare with is as suggested in \cite{Rasouli2019PIE}. They use a convolutional LSTM model on top of pre-trained VGG16 features obtained from cropped image regions of pedestrians using ground truth annotations (for a real life system, this input would come from the perception system and hence it is a sequential model) in 15 continuous frames. 



Our proposed method contains a multi-task architecture that consists of an object detection network called the primary network and a pedestrian intention network called auxiliary network as shown in Figure 2. For every input image, the auxiliary network receives its inputs from the feature maps calculated at an intermediate $L^{th}$ layer of the primary network and thus operates in parallel to the remainder of the primary network. Given a sequence of images from time T\textsubscript{0} to time T\textsubscript{t}, the primary network detects the bounding boxes of all the pedestrians in the images. The auxiliary network uses the information from previous frames and outputs the intention to cross for all the pedestrians in the frame at time T\textsubscript{t}. In our implementation, the primary network is a single shot YOLOv2 \cite{redmon2016yolo9000}  detector with output shape of $H \times W \times A \times (1 + N_C + 4)$, where A is the number of anchor boxes and $N_C$ defines the number of classes to be detected. We consider $N_C = 4$ with pedestrians, cross-walks, vehicles and traffic lights as the object classes since this can help the feature maps to capture relevant information regarding the spatial-temporal relationships between the scene objects and can be useful for the auxiliary pedestrian intention network. The auxiliary network is a three layered ConvLSTM model. It receives a time sequence of $t=15$ feature maps from an intermediate layer $L=18$ of the YOLOv2 and produces a similar spatial grid output of shape $H \times W \times A \times N_I$, where first three dimensions $H$, $W$ and $A$ has the same values as the primary network's output and the fourth dimension $N_I$ defines the number of pedestrian intention classes. The raw images of size $1920 \times 1080$ are resized to $640 \times 360$ before feeding as input to the YOLOv2 and our final layer parameters are $H=11$ and $W=20$ with $A=5$ anchor boxes.  $N_I= 2$ indicating whether the pedestrian intends to cross the street or not. We use binary cross entropy as the auxiliary loss. The last step is a constant time spatial mapping between the pedestrian detected from the primary network and the corresponding intentions from auxiliary network. If a pedestrian is detected at \{i,j,k\} grid in the YOLOv2 output, we assign the softmax output at the same \{i,j,k\} grid at the auxiliary network as the corresponding classified intention for the detected pedestrian. Thus, our architecture does not have to wait for the detection to be over before performing prediction. During the training process for the auxiliary side, only the loss corresponding to the cells that contain pedestrians are computed and back propagated to train the layers below. The modules can either be trained end-to-end in a multi-task learning manner or just the auxiliary network can be trained separately by using features from a pre-trained object detector. The latter scenario can be particularly useful when the perception network is either trained on a separate and comparatively bigger data set or in real life automotive system integration scenarios where the detection module may be obtained from a different party or supplier. For our experiments in the current paper, we assume an oracle tracking system and thus use the object tracking IDs directly from annotations. However, tracking can also be included in the whole end-to-end system, similar to as it is described in FaF \cite{luo2018fast}.

\begin{figure}[h]
  \centering
  \includegraphics[scale=0.40]{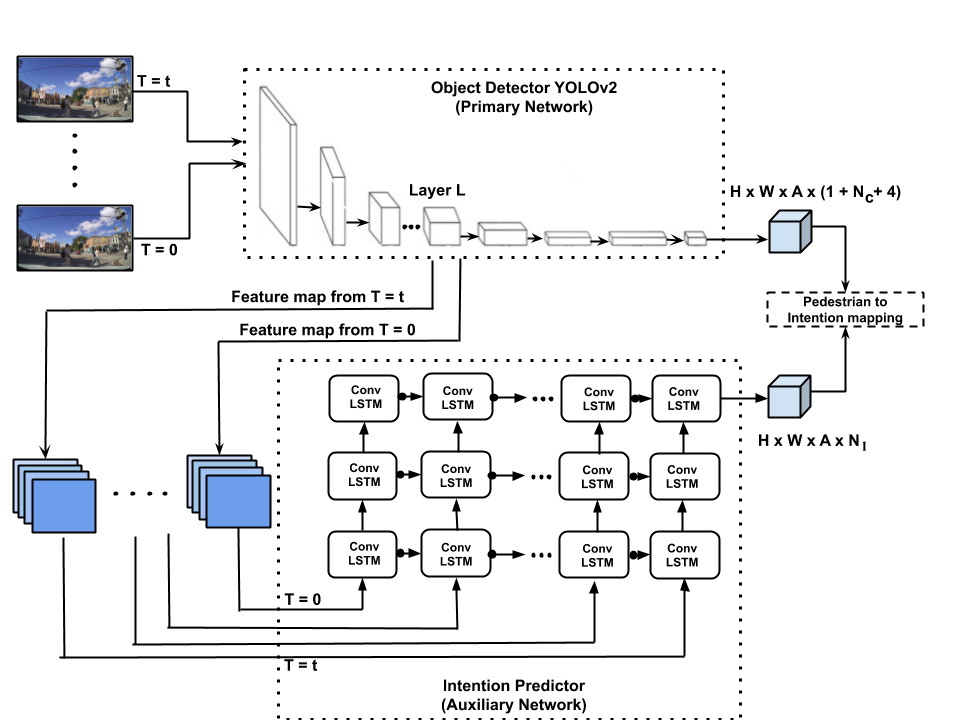}
  \caption{Spatial-temporal multitask network to do detection and intention prediction of all pedestrians simultaneously }
\end{figure}

\section{Experiments}
\subsection{Dataset}

In this paper, we use the PIE data set \cite{Rasouli2019PIE} to train both the detection and prediction models. The pedestrian intention was annotated using Amazon Mechanical Turks where each human subject was asked to observe a highlighted pedestrian in a sequence of consecutive frames  and answer whether the pedestrian wants to cross the street. Intention of a pedestrian is defined as the ultimate objective of the pedestrian and not the actual path the pedestrian executes based on the traffic scenarios. All videos are recorded in HD format (1920 $\times$ 1080 pixels) at 30 frames per second with intention annotated for a total of 1842 pedestrians. 

However, we found out that not all the pedestrians that can potentially interact with the ego vehicle are annotated. This posed difficulty for us in training the primary YOLOv2 network because of too many false negatives. Thus, in our current paper, we primarily compare the improvement in the intention prediction task only using our methodology over the sequential approach.

\begin{table}[t]
  \caption{Comparing intention prediction results for auxiliary and multitask approaches with baseline sequential approach. The branching is done at the $18^{th}$ layer of YOLOv2 detector}
  \label{result-table}
  \centering
  \begin{tabular}{c|c|c|c}
  \hline
  \textbf{Method} & \textbf{Pedestrian Height (px)} & \textbf{Accuracy} & \textbf{F1 Score} \\
  \hline
  \textbf{Sequential} & All peds $>$ 0 & 79\% & 87\%  \\
  \hline
  \multirow{2}{5em}{\textbf{Auxiliary Training}} & All peds $>$ 0 & 81.19\% & 88.67\% \\
  & All peds $>$ 120 & 82.00\% & 88.27\% \\
  \hline
  \multirow{2}{5em}{\textbf{Multitask Training}} & All peds $>$ 0 & 83.83\% & 90.59\% \\
  & All peds $>$ 120 & 84.37\% & 90.246\% \\
  \hline
  \end{tabular}
\end{table}

\begin{table}[t]
  \caption{Comparing intention predictions results for different values of $L$ }
  \label{SR-result-table}
  \centering
  \begin{tabular}{c|c|c}
  \hline
  \textbf{Input Layer} &  \textbf{Accuracy} &  \textbf{F1 Score}\\
  \hline
  \textbf{$17^{th}$} & 76.79\% & 83.85\% \\
  \hline
  \textbf{$19^{th}$} & 76.26\% & 83.87\% \\
  \hline
  \textbf{$20^{th}$} & 75.1\%  & 82.82\% \\
  \hline
  \end{tabular}
\end{table}

\begin{table}[t]
  \caption{Comparing memory and latency at the overall system level}
  \label{SR-result-table}
  \centering
  \begin{tabular}{c|c|c}
  \hline
  \textbf{Method} &  \textbf{Memory (MB)} &  \textbf{Time to prediction}\\
  \hline
  \textbf{Sequential} & 276.1 & 137.576ms \\
  \hline
  \textbf{Auxiliary/Multi-Tasking} & 261.5 & 50.4ms \\
  \hline
  \end{tabular}
\end{table}

\begin{figure}
\centering
\begin{minipage}{.45\textwidth}
  \centering
    \begin{subfigure}{\columnwidth}
         \centering
         \includegraphics[width=0.9\linewidth]{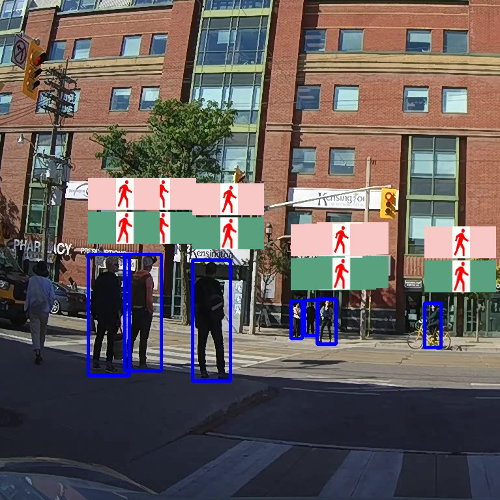}
         \caption{}
    \end{subfigure}
    \begin{subfigure}{\columnwidth}
         \centering
         \includegraphics[width=0.9\linewidth]{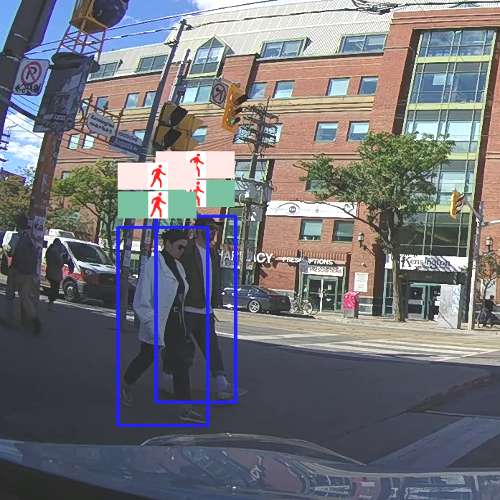}
         \caption{}
         \label{fig:three sin x}
     \end{subfigure}
    \begin{subfigure}{\columnwidth}
         \centering
         \includegraphics[width=0.9\linewidth]{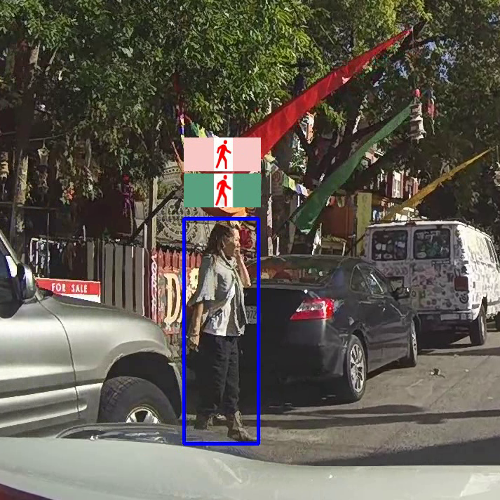}
         \caption{}
         \label{fig:five over x}
     \end{subfigure}
     \captionsetup{width=0.9\textwidth}
\caption{Pedestrian crossing results - Green background icon shows the ground truth intention and pink background icon shows the predicted intention}
  \label{fig:test1}
\end{minipage}
\begin{minipage}{.45\textwidth}
  \centering
  \begin{subfigure}{\columnwidth}
         \centering
         \includegraphics[width=0.9\linewidth]{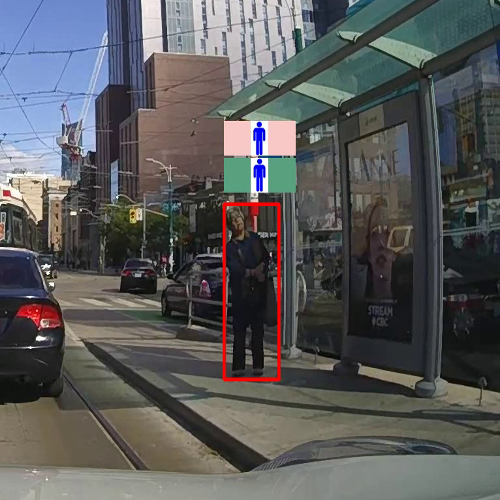}
         \caption{}
    \end{subfigure}
     \begin{subfigure}{\columnwidth}
         \centering
         \includegraphics[width=0.9\linewidth]{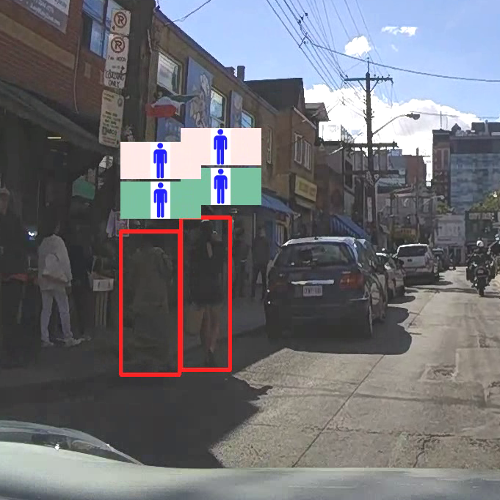}
         \caption{}
         \label{fig:three sin x}
     \end{subfigure}
     \begin{subfigure}{\columnwidth}
         \centering
         \includegraphics[width=0.9\linewidth]{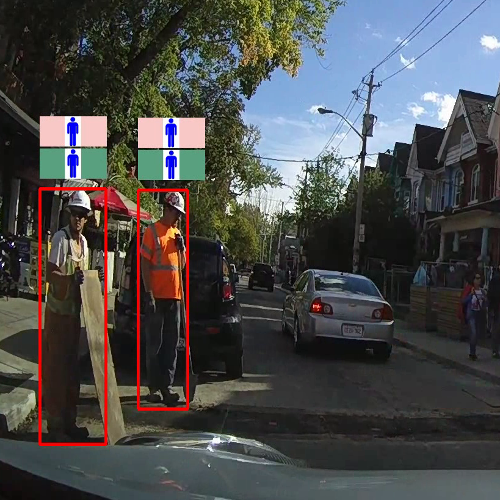}
         \caption{}
         \label{fig:five over x}
     \end{subfigure}
  \captionsetup{width=0.9\textwidth}
  \caption{Pedestrian not crossing- 
  Green background icon shows the ground truth intention and pink background icon shows the predicted intention}
  \label{fig:test2}
\end{minipage}
\end{figure}

\subsection{Quantitative Results}
The results shown in this section are on the default test set of the PIE data set. Table 1 summaries the accuracy and F1 score obtained for the baseline as well as the approaches presented in this paper for the $18^{th}$ layer feature map from YOLOv2 as input to auxiliary model. Using the code and model weights provided by \cite{Rasouli2019PIE}, we were able to reproduce the intention prediction accuracy of 79\% (and an F1 score of 87\% ) for the baseline sequential approach as claimed in their paper. 

For the auxiliary only training, we first trained the YOLOv2 object detector and subsequently froze all its layers. Given a training image, we pass the image through the frozen object detector, extract features from an intermediate layer and use that as input to the auxiliary intention network. For a sequence of images, we would then compute the loss for training only the layers in the auxiliary network. Keeping all the hyper-parameters same as set by \cite{Rasouli2019PIE} and trying out multiple different intermediate layers from object detector, the auxiliary approach was able to obtain an accuracy of 81.19\% and F1 score of 88.67\% with no size restriction on the pedestrians. Restricting the pedestrian size to a more humanly perceivable constraint ( that is by considering only those pedestrians with size more than 120 pixels in the original high definition whole scene image) increased the accuracy to 82\%. We found that the best accuracy is achieved when the  $18^{th}$ layer from YOLOv2 is used as the input to auxiliary network. We did a grid search to fix this hyperparameter and results from some other intermediate layers as input to auxiliary network are shown in Table 2. By $L^{th}$ layer, we mean taking the features generated after the application of batch normalization and activation on the $L^{th}$ convolutional layer of the YOLOv2 architecture.

Keeping all the hyper parameters same, we finally did the end-to-end multi-task training of both the detection and prediction networks simultaneously. This helped the common feature extraction layers in both detection and prediction tasks to be generic enough for both the tasks. This gave us a further increase in the accuracy to 83.83\% and F1 score to 90.59\%. Restricting the size of the pedestrian to a height of greater than 120 pixels only further increased the accuracy to 84.37\% and F1 score to 90.246\%. We thus see an improvement of 4.83\% in the pedestrian intention from the sequential approach. This end-to-end multi-task training also helped to improve the mean average precision (mAP) of the object detection model by 2\%.

Other major advantage of the proposed approach is the lower latency and lower parameter memory footprint. As shown in Table 3, the parameter memory of the auxiliary model was about 15 MB lower than the total memory of sequential approach. This can be attributed to the fact that the sequential method uses an extra feature extractor (VGG16) where as the auxiliary method utilizes the features extracted by the object detection network itself. The average inference time for a single image also reduced from 137.5 ms to 50.4 ms. This is because of the single shot inference behaviour of our approach in contrast to running the intention model for individual pedestrians separately in sequential method. Inference time of the sequential approach depends on the number of pedestrians in the image. Increased number of detected pedestrians increases the inference time for sequential approach. On the other hand, the inference time for the auxiliary approach is fixed and does not depend on the number of pedestrians in the image. The numbers reported here are corresponding to the average pedestrians per image in the PIE data set, which is 2.12. This would help for faster detection of the dynamic behaviour of the pedestrians. A more elaborate study on reducing the memory requirement and inference time with better hyper parameter optimization and architecture search is left as future work.

\subsection{Qualitative Results}
Figure 3 and 4 compare the predicted intention of the pedestrians with the ground truth annotations. In a crowded scene as seen in Figure 3(a), we can see the benefit of such a single shot approach for predicting the intention of all the pedestrians at the same time irrespective of the number of pedestrians in the image. Also, since the pedestrians are never cropped out of the original image, the intention network has access to the features of stimulus objects in the image such as traffic light, cross walk and vehicles. Figure 3(b) and 3(c) show the intention being predicted correctly for pedestrians that are walking and standing in the images respectively. Figure 4(a) shows a pedestrian who is waiting for something and the model correctly predicts the intention of not crossing. 
Figure 4(b) and (c) show pedestrians walking parallel to the ego vehicle and construction workers with no intention to cross the road respectively. Our model is able to accurately predict their intended behavior.

\section{Conclusion}



Given the complex dynamics and interactive nature of pedestrian motion patterns, it is extremely critical for autonomous vehicles to predict their intended behavior with maximum accuracy and in the most computationally efficient manner. We proposed a novel way to multitask detection and behavioral prediction directly from sensors data. We could show 2.7x inference speed increase and accuracy improvement of 4.83\%. It is worth noting here that the memory and latency improvement obtained by our methodology is orthogonal to other techniques like pruning, quantization and other hardware optimization of neural networks. Such multi-task methodology can be applied onto dealing with any combination of robotic perception and prediction tasks. For example, one can also use such spatial temporal feature sharing for single shot detection as well as prediction of the future trajectories of all vehicles in the scene in order to anticipate possible scenarios like cut ins and sudden braking. 

One possible weakness of our approach can be that the compressed features may or may not capture very fine details like hand gesture, eye gaze or head pose that may be needed to predict pedestrian intention whereas sequential approach may have an advantage here since its input is rectangular crops of pedestrians at the original whole image resolution. However, we hope that joint multi-task training can force the common layers to focus and retain such needed features. Although it requires further and extensive experiments to validate the safety and reliability of this methodology, our initial results seem encouraging to continue working in this direction.







\bibliography{main}

\end{document}